\documentclass[times,10pt,twocolumn]{article}
\usepackage{latex8}
\usepackage{times}
\usepackage{graphicx}
\usepackage{subfigure}
\usepackage{bm}
\usepackage{cite}
\usepackage{balance}
\usepackage[american]{babel}
\usepackage{paralist,algorithmic,algorithm}
\usepackage{amsmath,mathrsfs,amssymb}
\usepackage{color}
\def \D {\mathcal{D}}
\def \H {\mathcal{H}}
\def \S {\mathcal{S}}
\def \R {\mathbb{R}}

\newtheorem{myDef}{Definition}

\begin{document}
\title{Learnability with Time-Sharing Computational Resource Concerns}

\author{Zhi-Hua Zhou\\
National Key Laboratory for Novel Software Technology\\ Nanjing University, Nanjing 210023, China\\
zhouzh@nju.edu.cn}

\maketitle

Conventional machine learning theories generally assume explicitly or implicitly that there are enough or even infinitely supplied computational resource such that \textit{all received data can be handled}. In real practice, however, this is not the case. For example, in \textit{stream learning} the incoming data streams can be potentially endless with overwhelming size and it is impractical to assume that all received data can be handled in time. Indeed, the performance of machine learning depends not only on how many data have been received, but also on how many data can be handled subject to computational resource available; this is beyond the consideration of conventional learning theories.

Current ``intelligent supercomputing'' facilities generally work in an \textit{exclusive} way: a user is allocated a pre-set amount of resources to run her machine learning task. Because the amount is pre-set, it can be too optimistic such that the task could not complete, or too pessimistic such that fewer resources are really needed and some resources should have been allocated to other users. This looks like early computer systems that were only able to serve a single user program. With great effort of computer science pioneers, our computer systems are able to provide \textit{reasonable} service to each program, where the key technique is \textit{time-sharing}.

Time-sharing has two meanings according to Turing award laureate Fernando J. Corbat\'{o} \cite{Corbato62}. One is concerned with \textit{user efficiency}, trying to help user to get a fast response from the system. The other is concerned with \textit{hardware efficiency}. As explained by Turing award laureate Edgar F. Codd \cite{Codd60}: ``\textit{in order to exploit fully a fast computer ... the construction of a schedule entails determining which programs are to be run concurrently and which sequentially with respect to each other ... tends to minimize the time for executing the entire pending workload, subject to external constraints such as precedence, urgency, etc.}'', where a \textit{scheduling} mechanism  executes each program in some order, for some time, not necessarily to completion.

We believe that the concerns of time-sharing computational resources should be taken into account in machine learning theories. On one hand, users wish to get the result of training a satisfactory model within certain time budget; this corresponds to user efficiency. On the other hand, computational resource should be wisely exploited; this corresponds to hardware efficiency. A learning theory with time-sharing computational resource concerns won't assume that all received data can be handled in time, where scheduling is crucial.

For this purpose, we define \textbf{Computational Resource Efficient Learning} (CoRE-Learning) and present a theoretical framework.

First, we introduce the notion of \textit{machine learning throughput}. Throughput is a basic concept in computer networking, defined as the amount of data per second that can be transferred~\cite{KuroseRoss16}; it is also concerned in database systems to measure the average number of transactions completed within a given time~\cite{book:database}. The introduction of throughput enables us to theoretically formulate the influence of computational resource and scheduling at an abstract level.

\begin{figure}[!t]
\begin{center}
  \includegraphics[width=.85\linewidth]{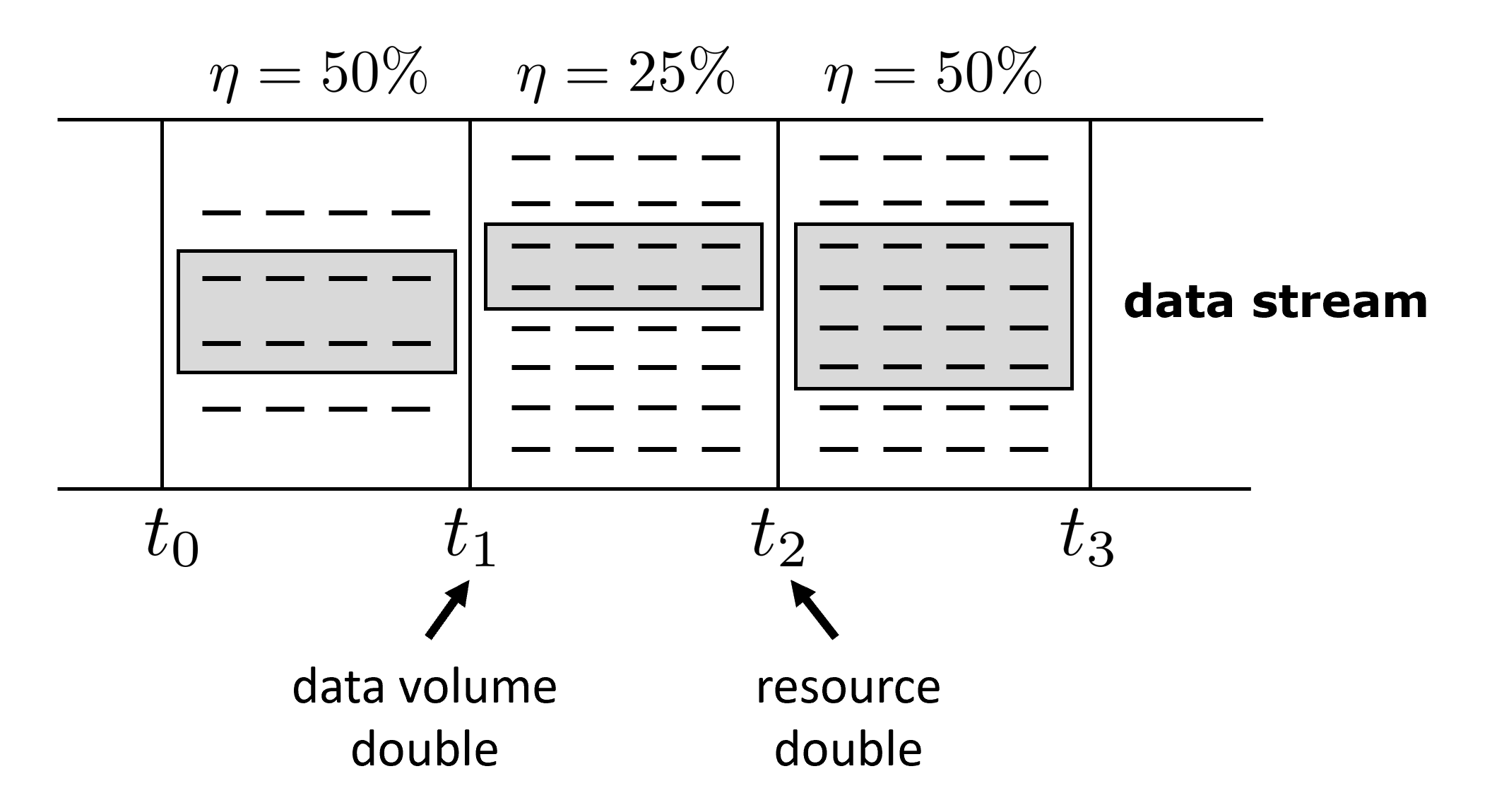}\vspace{-3mm}
  \caption{Illustration of data throughput.}\label{fig:fig1}
\end{center}
\end{figure}

Our proposed machine learning throughput involves two components. The first component is \textit{data throughput}. As illustrated in Figure~\ref{fig:fig1}, data throughput represents \textit{the percentage of data that can be learned per time unit}. For example, half of the received data can be timely exploited in the time unit $t_0 \sim t_1$ in Figure~\ref{fig:fig1}, corresponding to a data throughput $\eta = 50\%$. In the time unit $t_1 \sim t_2$, the data volume doubles such that only $25\%$ of received data can be timely exploited with current resource, and thus, $\eta$ becomes $25\%$. While in the time unit $t_2 \sim t_3$, the resource doubles such that $\eta$ becomes $50\%$ again. It is evident that the influence of data volume as well as the computational resource budget can be involved by introducing the notion of data throughput into machine learning studies. The above discussion does not take into account that the difficulty of learning from the data may vary since unknown changes may occur; this is related to \textit{open-environment machine learning}~\cite{Zhou22} and the consideration can be accommodated in further studies.

We call a machine learning task received by the supercomputing facility as a \textit{thread}. It is associated with two time points: a beginning time and a deadline time, specifying lifespan of the thread. If the thread can be well learned (i.e., the performance reaches user's demands) within its timespan, we call it a \textit{successful thread}, and otherwise a \textit{failure thread}. Note that if we set the deadline time according to user's learning rapidity requirement about the thread, then a thread is successful if a satisfactory model can be learned within our given time budget.

\begin{figure}[!ht]
\begin{center}
  \includegraphics[width=.85\linewidth]{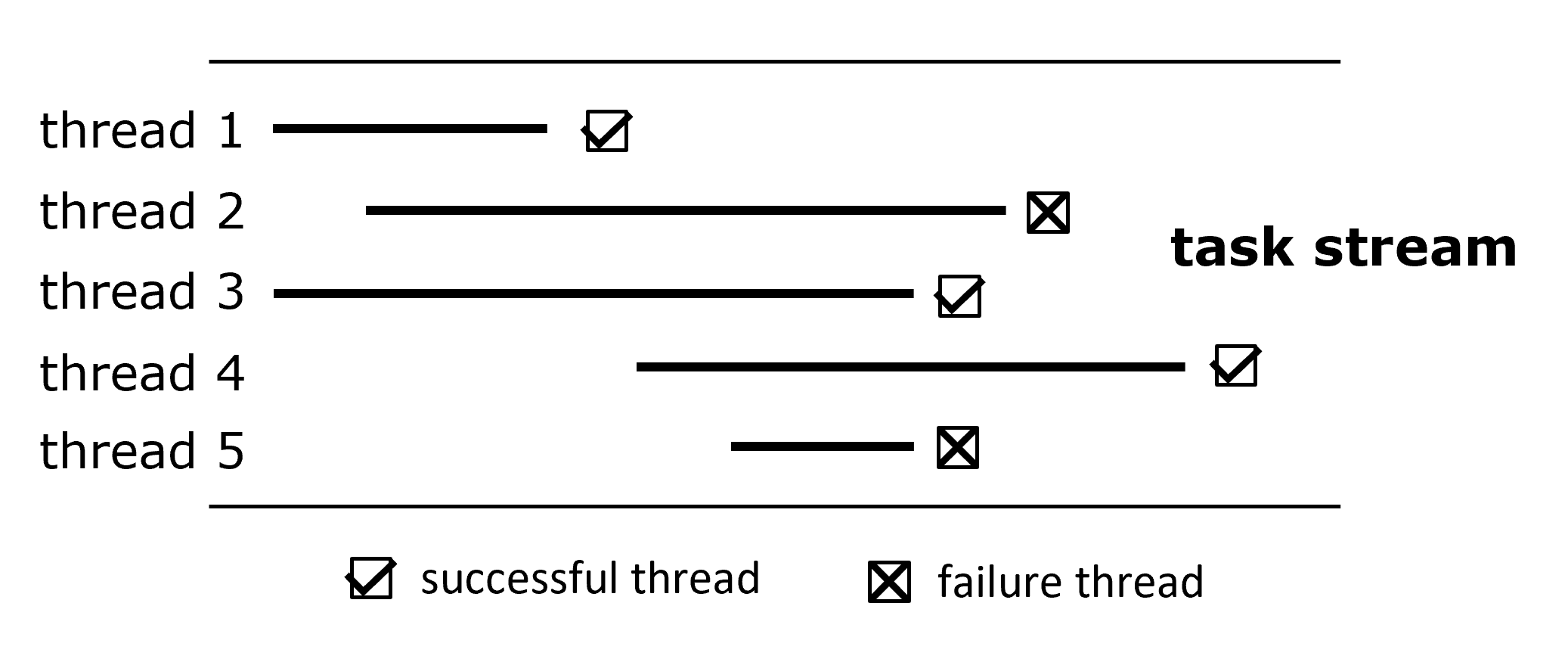}
  \caption{Illustration of thread throughput.}\label{fig:fig2}
\end{center}
\end{figure}

Now, we introduce the second component of machine learning throughput, i.e., \textit{thread throughput}, defined as \textit{the percentage of threads that can be learned well in a time period}, calculated by the percentage of successful threads in all threads during that time period. As illustrated in Figure~\ref{fig:fig2}, the thread throughput is $\kappa = 60\%$.

Let $\S = \left(\{\mathcal{T}_k\}_{k=1}^K, \{N_t\}_{t=1}^T\right)$ denote a \textit{task bundle}, i.e., a set of task threads during the concerned time period, where $\mathcal{T}_k = (\D_k, b_k, d_k)$ denotes the $k$-th thread with data distribution $\D_k$, beginning time $b_k$ and deadline time $d_k$. $N_t$ is the amount of data that can be handled at time $t$ given the total budget of computational resource. $K$ is the total number of threads in the task bundle, and $T$ is the total number of timeslots. Note that if $b_i = b_j ~(\forall i \neq j)$, then all task threads arrive at the same time.

A learning algorithm $\mathcal{L}$ receives $\S$ as input. $\mathcal{L}$ will output $\{(s_k, M_k)\}_{k=1}^{K}$, where $s_k$ is the switching time determined by the algorithm and $M_k$ is the learned model for the $k$-th thread. We use $A_t$ to denote the set of alive threads, i.e., $A_t = \{k \mid b_k\leq t \leq d_k \mbox{ and } k \in [K]\}$. The learning process proceeds as follows.

\smallskip

\begin{algorithmic}[1]
  \FOR{time $t = 1, \ldots, T$, the learner }
    \STATE collects at most $\eta_{k,t} N_t$ samples for thread $k \in A_t$, where $\eta_{k,t}$ is data throughput for thread $k$ at time $t$;
    \STATE updates the model $M_k$ for thread $k$;
    \STATE if thread $k$ completes, set $s_k \leftarrow t$;
  \ENDFOR
\end{algorithmic}

\smallskip

Now we introduce \textbf{CoRE-learnability}, with $\eta$ and $\kappa$ denoting data throughput and thread throughput, respectively.

\begin{myDef} [{$(\eta,\kappa, \mathcal{L})$-CoRE learnability}] A task bundle $\S = (\{\mathcal{T}_k\}_{k=1}^K, \{N_t\}_{t=1}^T)$ is $(\eta,\kappa, \mathcal{L})$-CoRE learnable, if there exists a computational resource scheduling strategy $\psi$ that enables $\mathcal{L}$ to output $\{(s_k, M_k)\}_{k=1}^{K}$ running in polynomial time in $1/\epsilon$ and $1/\delta$ such that for some small $\epsilon$ and $\delta$, with probability at least $1 - \delta$,
\begin{enumerate}
  \item[(1)] for all $t \in [T]$, $\sum_{k \in A_t} \eta_{k,t} \leq \eta$;
  \item[(2)] $|I_{\text{succ}}| \geq \kappa K$,
    \begin{enumerate}
    \item[(2a)] $s_k \leq d_k$, for all $k \in I_{\text{succ}~}$,
    \item[(2b)] $R_k(M_k) \leq \epsilon$, for all $k \in I_{\text{succ}~}$,
    \end{enumerate}
\end{enumerate}
where $I_{\text{succ}}$ (or $I_{\text{fail}}$) is the set of successful (or failure) threads, $I_{\text{succ}} \cap  I_{\text{fail}} = \emptyset$, $I_{\text{succ}} \cup I_{\text{fail}} = [K]$.
\end{myDef}

Condition (1) concerns data throughout, constraining that the overall resource quota of threads in the alive set never exceeds the maximum resource budget. Condition (2) concerns thread throughput, demanding the scheduling strategy $\psi$ to enable $\mathcal{L}$ to learn as many threads well as possible: the learning of the thread should be completed before the deadline, as indicated by condition (2a); and the learning performance of the thread should be within a small error level, as indicated by condition (2b). The learning performance is measured by $R_k: \H_k \mapsto \R$, and $R_k(M_k) \leq \epsilon$ evaluates whether the learning performance is acceptable according to a predetermined $\epsilon$ when the algorithm exploits data received in the timeslot $(b_k, s_k)$ and completes learning by the time point $s_k$. Note that Condition (1) is related to user efficiency, while Condition (2) is related to hardware efficiency; the scheduling strategy should balance the two aspects carefully.

The CoRE-learnability definition employs an $(\epsilon,\delta)$-language similar to the PAC (Probably Approximately Correct) learning theory~\cite{STOC'84:Valiant-PAC}. It is noteworthy that, however, PAC learning theory focuses on learning from data sampled from an underlying data distribution, assuming that all training data can be exploited in time; thus, it allows for an arbitrarily small error $\epsilon$ and an arbitrarily high confidence $1-\delta$ given that the number of samples is sufficiently large (but still can be well exploited in time). In contrast, CoRE-learning theory considers the influence of the resource scheduling strategy $\psi$, and demands only acceptable $(\epsilon, \delta)$ for $\mathcal{L}$ with $(\eta, \kappa)$ throughput concerns.

\begin{figure}[!t]
\begin{center}
  \includegraphics[width=\linewidth]{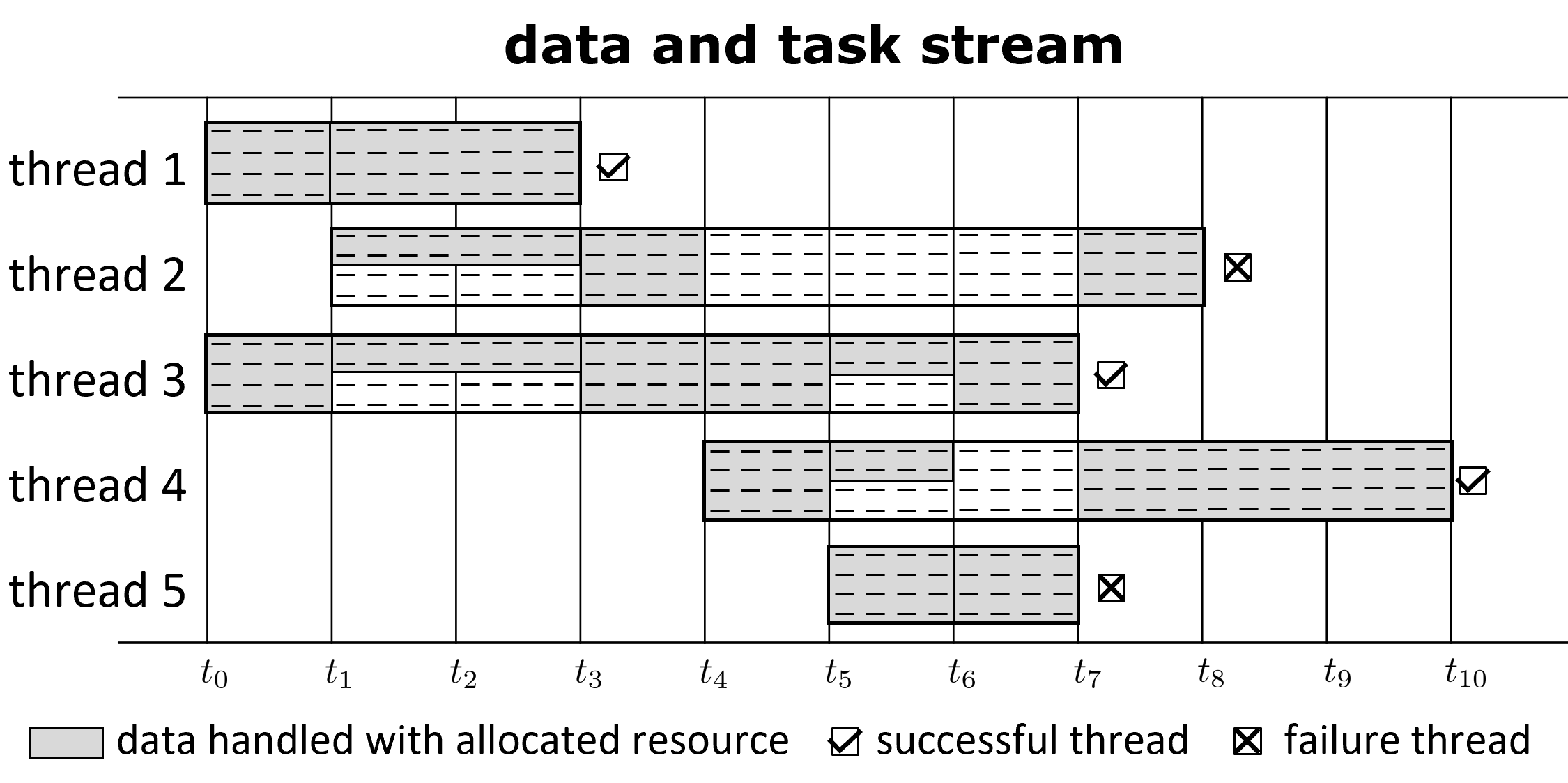}
  \caption{An illustration of CoRE-learnability.}\label{fig:fig3}
\end{center}
\end{figure}

Figure~\ref{fig:fig3} presents an illustration, where the task bundle consists of $K = 5$ threads. For simplicity, assume that in each time unit $N_t = N = 64$ data units can be handled. Note that CoRE-learning allows the beginning time $b_k$ and deadline time $d_k$ of the task thread $\mathcal{T}_k = (\D_k, b_k, d_k)$ to be any real value, while in this figure we assume that they are rounded up for a better illustration. For a given algorithm $\mathcal{L}$,
the task bundle is $(.5,.6, \mathcal{L})$-CoRE learnable, because there exists a scheduling strategy $\psi$ that enables $\mathcal{L}$ to successfully learn three out of the total five threads given a data throughput $\eta = 50\%$. As Figure~\ref{fig:fig3} shows, $\psi$ allocates resource that can handle $\eta N = 32$ data units equally to threads 1 and 3 in $t_0 \sim t_1$. Thread 1 continues to get resource that can handle 16 data units until it completes at $t_3$; the remaining resource that can handle 16 data units are allocated to threads 2 and 3 equally in $t_1 \sim t_3$. In $t_3 \sim t_4$ threads 2 and 3 each receives resource that can handle 8 more data units because thread 1 does not require resource anymore. Thread 4 comes at $t_4$, while $\psi$ decides to allocate all resource to threads 3 and 4 as it feels pessimistic about thread 2. At $t_5$, thread 5 comes, and because its lifespan is quite short, $\psi$ decides to allocate it as many resource as possible, until the learning of thread 5 fails at $t_7$. At $t_6$, $\psi$ feels very optimistic about thread 3, and therefore, it decides to give it all remaining resource, at the cost of sacrificing thread 4 temporarily. At $t_7$ there are only threads 2 and 4 alive. Finally, threads 2 and 5 fail for different reasons: thread 2 fails because of unsatisfactory learning performance, violating condition (2b), whereas thread 5 fails to complete before the deadline, violating condition (2a).

\begin{figure}[!t]
\begin{center}
  \includegraphics[width=.95\linewidth]{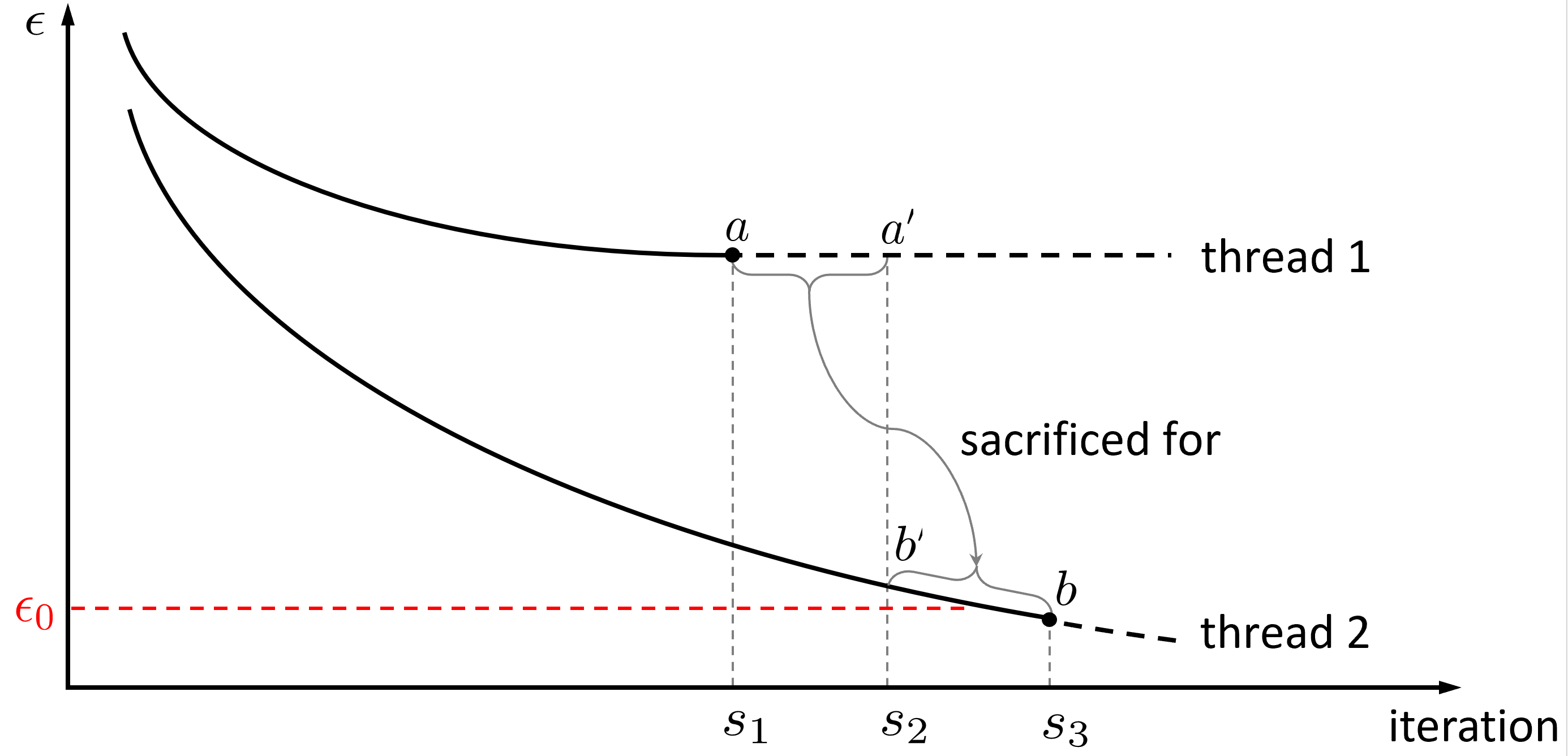}\vspace{-3mm}
  \caption{An illustration of adaptive resource allocation.}\label{fig:fig4}\vspace{-3mm}
\end{center}
\end{figure}

The resource scheduling strategy $\psi$ is able to allocate resource adaptively, based on perceiving the learning status and foreseeing the learning progress of the threads. Intuitively, if $\mathcal{L}$ is based on gradient calculation, then the allocation of more computational resource to a task implies that more gradient calculations can be executed for that task. As Figure~\ref{fig:fig4} illustrates, assume that the two task threads are allocated with the same amount of resource initially. At iteration $\tau_1$, $\psi$ perceives that thread 1 arrives at a flat convergence area where its error has not significantly dropped during the past five rounds of gradient calculation, whereas thread 2 goes into a slope area with a faster error drop. Then, $\psi$ decides to reduce the resource for thread 1 and reallocates them to thread 2. At the final iteration $\tau_3$, thread 2 reaches the status $b$ rather than $b'$, with the sacrifice of thread 1 which reaches the status $a$ rather than $a'$, leading to a better overall throughput of $.5$ (i.e., thread 2 is judged to be successful according to the threshold $\epsilon_0$) rather than $.0$ (i.e., neither threads reach $\epsilon_0$ if the computational resource continue to be evenly allocated). Indeed, even if one considers another definition for thread throughput, such as defining it according to average error, the helpfulness of $\psi$ is still visible from the improvement from $\frac{\epsilon_{a'} + \epsilon_{b'}}{2}$ to $\frac{\epsilon_{a} + \epsilon_{b}}{2}$.
Merely maximizing thread throughput may lead $\psi$ to prefer learning easier threads; this can be repaired by assigning priority or importance weights to threads when needed.

The CoRE-learning discussed in this article enables the influence and scheduling of computational resources be taken into account in learning theory. One of the fundamental goals is to, by introducing \textit{scheduling}, enable the computational resource for machine learning to be used in a \textit{time-sharing} style rather than the current \textit{elusive} style. For example, even though the \textit{scaling law} in training large language models is well-known, resources used to train such models are still used in an elusive way, leading to big waste because it is hard to pre-set a just-right amount. Distributed machine learning \cite{book:Liu} tries to partition a learning task for distributed computing, where at each distributed site the resource is still exploited in an elusive way with a pre-set amount of resources, and the focus is on how to minimize the communication cost and guarantee the convergence by adequately synchronizing calculations.

Note that resource scheduling in machine learning is very different from that in other fields such as computer systems and databases. For example, the amount of resources required for accomplishing a task in computer systems and databases is generally known once the task is received, whereas in machine learning this information is unknown and can only be estimated by spying on the learning process online. This raises new research issues that might have been overlooked before, such as how to govern a machine learning process and estimate its status and progress \textit{online} effectively and efficiently. It is even more complicated when noticing that the online governing and status estimation require communication and computational resources. Thus, CoRE-learning naturally involves an exploration-exploitation balance with resource scheduling. CoRE-learnability of concrete CoRE-learning algorithms can be proved once such algorithms are developed.

\bibliographystyle{plain}
\bibliography{bibstream}
\end{document}